\documentclass[sigconf]{acmart}

\makeatletter
\renewcommand\@formatdoi[1]{\ignorespaces}
\makeatother

\AtBeginDocument{%
  \providecommand\BibTeX{{%
    \normalfont B\kern-0.5em{\scshape i\kern-0.25em b}\kern-0.8em\TeX}}}

\setcopyright{none}
\renewcommand\footnotetextcopyrightpermission[1]{}

\setcopyright{acmcopyright}
\copyrightyear{2021}
\acmYear{2021}
\acmDOI{10.1145/1122445.1122456}

\acmConference[KDD 2021]{City Brain Challenge @ KDD 2021 Workshop}{August 14-18, 2021}{Singapore}
\acmPrice{15.00}
\acmISBN{978-1-4503-XXXX-X/18/06}

\begin{document}

\title{DQN Control Solution for KDD Cup 2021 City Brain Challenge}

\author{Yitian Chen}
\email{yitianartsky@gmail.com}
\affiliation{%
  \institution{ BIGO Technology}
  \country{}
}

\author{Kunlong Chen}
\email{chenkunlong@bigo.sg}
\affiliation{%
  \institution{ BIGO Technology}
  \country{}
}

\author{Kunjin Chen}
\email{kunjin.ckj@alibaba-inc.com}
\affiliation{%
  \institution{Alibaba Group}
  \country{}
}

\author{Lin Wang}
\email{marvin.wl@alibaba-inc.com}
\affiliation{%
  \institution{Alibaba Group}
  \country{}
}

\begin{abstract}
    We took part in the city brain challenge competition\footnote{\url{http://www.yunqiacademy.org/poster}} and achieved the 8th place.
    In this competition, the players are provided with a real-world city-scale road network and its traffic demand derived from real traffic data. The players are asked to coordinate the traffic signals with a self-designed agent to maximize the number of vehicles served while maintaining an acceptable delay.
    In this abstract paper, we present an overall analysis and our detailed solution to this competition. Our approach is mainly based on the adaptation of the deep Q-network (DQN) for real-time traffic signal control. 
    From our perspective, the major challenge of this competition is how to extend the classical DQN framework to traffic signals control in real-world complex road network and traffic flow situation. 
    After trying and implementing several classical reward functions~\cite{abdoos2011traffic, wei2019presslight}, we finally chose to apply our newly-designed reward in our agent.
    By applying our newly-proposed reward function and carefully tuning the control scheme, an agent based on a single DQN model can rank among the top 15 teams.
    We hope this paper could serve, to some extent, as a baseline solution to traffic signal control of real-world road network and inspire further attempts and researches.
\end{abstract}



\maketitle

\section{Introduction}
Traffic signals coordinate the traffic movements at the intersections and a smart traffic signal coordination algorithm is the key to transportation efficiency.
In this competition, the players are provided with a real-world city-scale road network and its traffic demand derived from real traffic data.     
The players are asked to coordinate the traffic signals with an self-design agent.
For each period of time step in an intersection, the agent can select one of 8 types of signal phases, serving a pair of non-conflict traffic movements (e.g., phase-1 gives right-of-way for left-turn traffic from northern and southern approaches), to maximize the number of vehicles served while maintaining an acceptable delay.

\begin{figure}[htp!]
  \centering
    \includegraphics[width=0.5\textwidth]{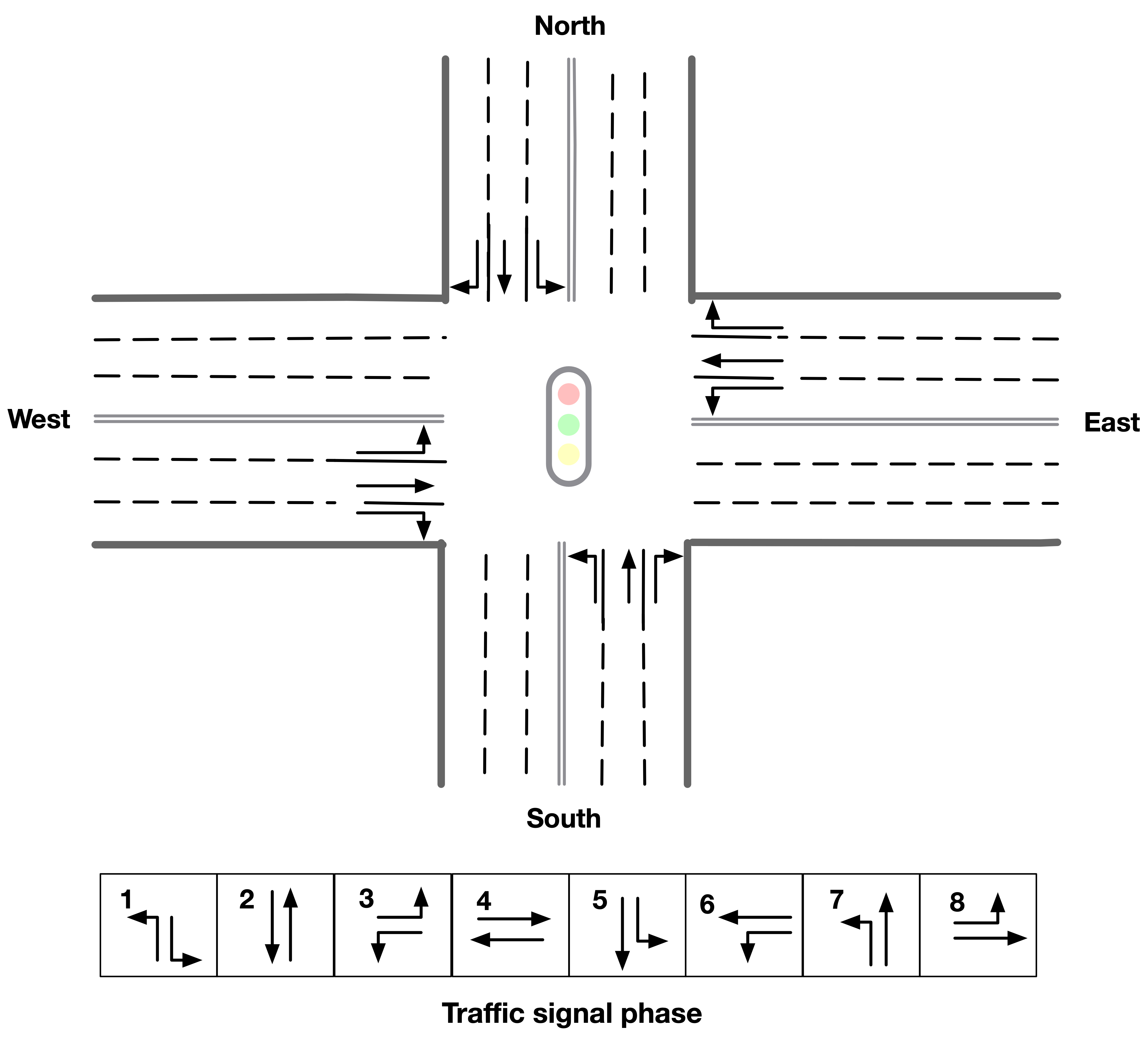}
    \caption{An example of a four-leg intersection and its phase types.}
    \label{fig:intersection}
\end{figure}
Figure~\ref{fig:intersection} shows an example of four-leg intersections and 8 types of signal phases.
The evaluation metrics used in this competition is the total number of vehicles served.
In the final phase, the delay index is computed every 20 seconds to evaluate the players' final score.
The evaluation process is terminated once the delay index reaches the predefined threshold , which is set to 1.40.

As introduced by the competition organizing committee, lots of reinforcement learning based (RL-based) approaches have been proposed and achieved state-of-the-art result in simulation environment~\cite{wei2019presslight,abdoos2011traffic,wei2019survey,zheng2019learning}.
However, most of researches are based on the simulation environment and there is little work derived from real-world city-scale road network and traffic data. 
This competition provides a good opportunity to valid the performance of different RL-based approaches in real-world situation.
In this paper, we present an overall analysis of the competition and our DQN-based solution framework to settle the city brain challenging.

\section{Preliminaries}
\begin{table*}[!h]
        \centering
        \caption{Summary of notations}
        \begin{tabular}{ll}
                \toprule
                Notation & Meaning    \\ \hline
                $v$  & A vehicle in the traffic flow. \\
                $d(v)$ & Delay index of vehicle $v$, $d(v)= speed(v)/speedLimit(v)$.   \\
                $q(v)$ & Queue status of vehicle $v$, $q(v) = Bool(speed(v)<0.3)$ \\
                $I$ &  Set of intersections. \\
                $J$ & Lane index of an intersection, ranging from 0 to 23. \\

                $A_i^k$ & Zone of influence of intersection $i$ with distance $k$. \\
                $x^t_j(A_i^k)$ & Vehicle number on lane index $j$ of zone of influence $A_i^k$ of intersection $i$ at step $t$. \\
                $d^t_j(A_i^k)$ & Delay index on lane index $j$ of zone of influence $A_i^k$ of intersection $i$. at step $t$  \\
                $q^t_j(A_i^k)$ & Queue length on lane index $j$ of zone of influence $A_i^k$ of intersection $i$ at step $t$. \\
                \bottomrule
        \end{tabular}
        \label{tab:xl}
\end{table*}
\subsection{Environment and Data Description}

\begin{figure}[htp!]
  \centering
    \includegraphics[width=0.45\textwidth]{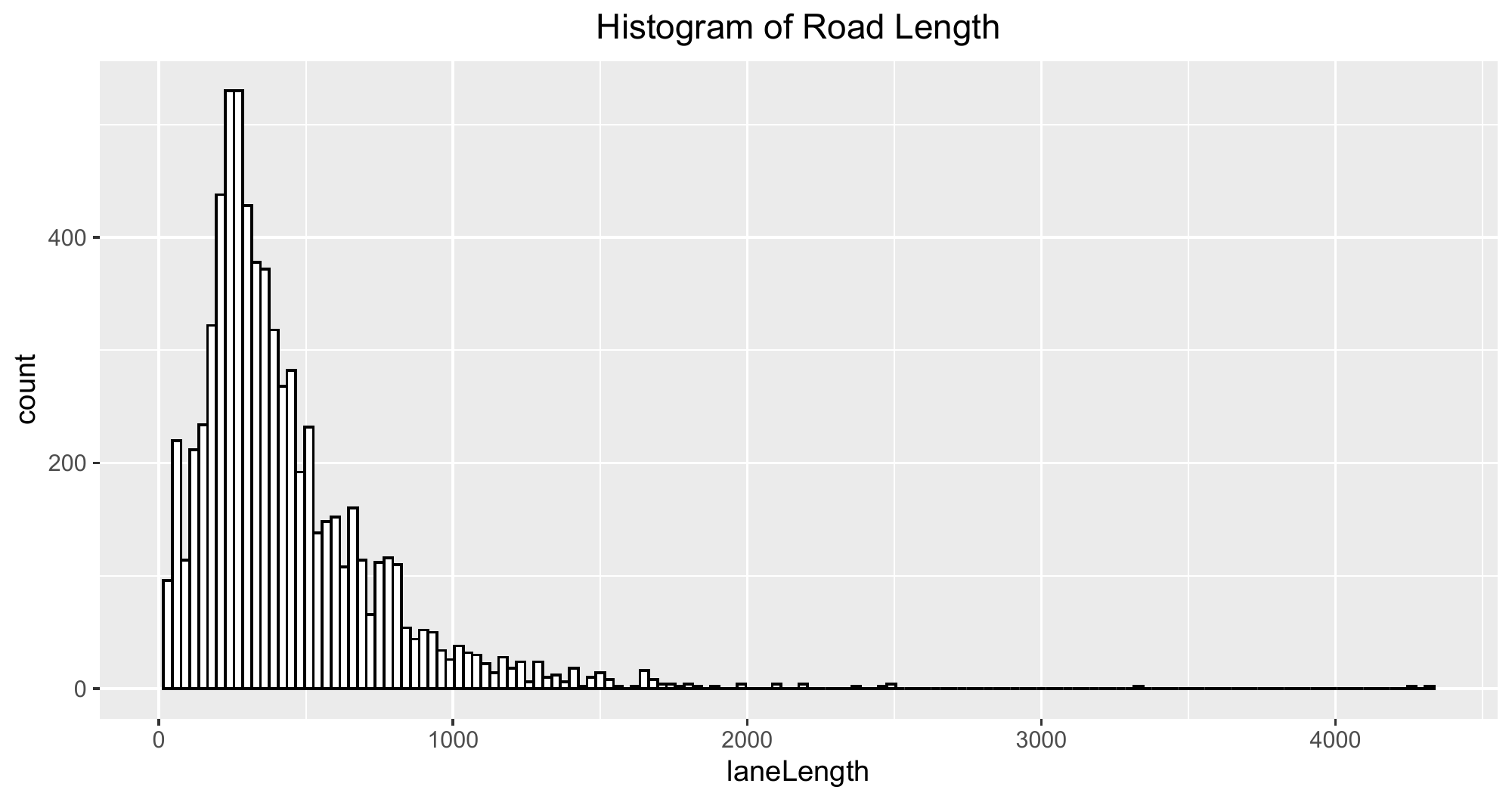}
    \caption{The distribution of road length in the competition.}
    \label{fig:dist}
\end{figure}

Figure~\ref{fig:dist} shows the distribution of road lengths in this competition, which ranges from 37 meters to more than 4,000 meters.
Meanwhile, an action of traffic signal phase to be selected will last for 10 seconds in the default setting. As a result, it may be ineffective to take all vehicles in the corresponding lanes of the intersection into consideration.
Our first work is to define the \textit{zone of influence}. Specifically, for each intersection $i$, we define the segment of each lane where the distance to its center is less than $k$ meters as the zone of influence and denote it as $A_i^k$. 
Meanwhile, for each lane, the lane length is set as the distance capping of $k$, which can be expressed as:
\begin{equation*}
k := \min(k, \text{lane length})
\end{equation*}
The design and calculation of the states and rewards in our solution framework hereinafter are based on this definition.

\subsection{Notations and Definition}

In Table~\ref{tab:xl}, we summarize some important notations used in our framework.
For a vehicle $v$ extracted from the objective ``info'' in simulation environment , we define the delay of the vehicle as 1 minus the speed $v$ dividing the speed limit of the current lane $speedLimit(v)$ 
\begin{equation}
    d(v) = 1 - \frac{speed(v)}{speedLimit(v)}
\end{equation}
Meanwhile, we define the queue status $q(v)$ to indicate whether a vehicle has a speed smaller than $0.3$ m/s. $J$ is the index set of the lanes of an intersection, ranging from 0 to 23. The index of a nonexistent lane is marked as $-1$. We further define $x^t_j(A_i^k)$,$d^t_j(A_i^k)$ ,$q^t_j(A_i^k)$ as the number of vehicles, the delay index, and the queue length within the zone of influence $A_i^k$ on lane $j$ at step $t$, respectively. More specifically, the delay index of a lane is the average delay of all vehicles being considered. For a given intersection $i$ and distance $k$, one can easily obtain the statistics for each lane based on the information of vehicles within the zone of influence at time step $t$.

In addition, the definition of the actions are clearly described in the official document:
 at step $t$, for each intersection $i$, the agent chooses a phase $p_{it}$, indicating that the traffic signal should be set to phase $p_{it}$, which is an element in $\mathcal{S} =\{1,2,3,4,5,6,7,8\}$.

\section{Method}
In this section, we first present the state and reward design for the agent.
Then we illustrate the control schemes, model design, and model training schemes of the proposed DQN framework.
Finally, we describe our rule-based agent. 
As we will show in Section~\ref{sec:experiment}, an ensemble of DQN-based agent and rule-based agent can achieve out best evaluation score on the leader-board. 

\subsection{State Design}
For each intersection, the state includes the current phase $p$, the current time step, the time duration of current phase and the statistic features of vehicle number, delay index, queue length and pressure for each phase based on the definition of zone of the influence.

Table~\ref{tab:state} shows the statistic features we use in the state design.
To calculate the aggregate feature of each traffic signal phase, we extract the feature for the corresponding lanes given a specific value of $k$ and add them together.

For instance, for the pair of feature ``vehicle number'' and $k=60$ of signal phase 1, we summarize the total vehicle number in the segments of of lane $0$ and $6$ within zone of influence with $k=60$, and it can be expressed as:
\begin{equation*}
x^t_0(A_i^{60})+ x^t_6(A_i^{60})
\end{equation*}
The same state for other phases can be obtained likewise. As the number of total signal phases is 8, correspondingly, the number of feature dimension is 8.

Further, for the pair of feature ``pressure of vehicle number'' and $k=60$ of signal phase 1, in addition to upstream vehicle counts, we summarize the total downstream vehicle numbers on lane list $[15, 16, 17, 21, 22, 23]$ within zone of influence with $k=60$ and divide it by 3.
To calculate the pressure, we use the upstream vehicle number minus the downstream vehicle number divided by 3, which can be express as:
\begin{equation}
\begin{split}
x^t_0(A_i^{60})+ x^t_6(A_i^{60})- &\frac{1}{3} [x^t_{15}(A_i^{60})+ x^t_{16}(A_i^{60})+x^t_{17}(A_i^{60})\\ &+ x^t_{21}(A_i^{60})
+ x^t_{22}(A_i^{60})+x^t_{23}(A_i^{60}) ]
\end{split}
\end{equation}

\begin{table}[htp!]
        \centering
        \caption{Statistic features used in the state design. For each pair , e.g., the feature vehicle number and zone of influence with distance $k=60$ , the feature dimension is 8, corresponding to 8 signal phases.}
        \begin{tabular}{lccc}
                \toprule
                & \multicolumn{3}{c}{Feature dimension} \\  \cline{2-4}
            Feature group& $k=60$ & $k=100$  & $k=200$\\
                \midrule
            Vehicle number & 8 & 8 & 8\\
            Delay index & 8 & 8 & 8 \\
            Queue length & 8 & 8 & 8 \\
            Pressure of vehicle number  & 8 & 8 & 8 \\
            Pressure of delay index & 8 & 8 & 8 \\
            Pressure of queue length  & 8 & 8 & 8 \\
                \bottomrule
        \end{tabular}
        \label{tab:state}
\end{table}

\subsection{Reward Design}
        Various reward designs have been proposed in the literature.
        Our first attempt is to choose the rewards of queue length and delay~\cite{wei2019survey}.
        Specifically, given an intersection $i \in I$ at time $t$ and its' zone of influence $A_i^{k}$, the rewards of queue length or delay index at time $t$ is the sum of all queue length or delay index in the next timestap $t+1$.
        \begin{equation}
            \begin{aligned}
                R^{\rm{delay}}(t) & = - \sum\limits_{j=0}^{23} (d_j^{t+1}(A_i^k))\\
                R^{\rm{queue}}(t) & = - \sum\limits_{j=0}^{23} (q_j^{t+1}(A_i^k))
            \end{aligned}
        \end{equation}
        A combination of delay index and queue length (we denote it as ``DQ'') can empirically achieve better results:
        \begin{equation}
            R^{\rm{DQ}}(t) =  - \sum\limits_{j=0}^{23} (d_j^{t+1}(A_i^k)+q_j^{t+1}(A_i^k))
        \end{equation}
        Another reward recommended by the official competition organizers is max pressure (MP)~\cite{wei2019presslight}. 
        For an intersection $i$ of time $t$, the MP reward is defined as the sum of the absolute pressures over all traffic movements:
        \begin{equation}
            R^{\rm{MP}}(t)  =  \left| \sum\limits_{j=0}^{11} x_j^{t+1}(A_i^k) - \sum\limits_{j=12}^{23} x_j^{t+1}(A_i^k)  \right|
        \end{equation}
        However, the performance of the original reward function is relatively poor. In our implementation, combining the MP reward with the DQ reward can achieve better results:
         \begin{equation}
         \begin{aligned}
             R^{\rm{MP-DQ}}(t)  = & - \sum\limits_{j=0}^{11} (d_j^{t+1}(A_i^k)+q_j^{t+1}(A_i^k)) \\ & + 1/2 \sum\limits_{j=12}^{23} (d_j^{t+1}(A_i^k)+q_j^{t+1}(A_i^k)) 
        \end{aligned}
        \end{equation}
        Further, the reward function that achieves the best single-model result is a new one we proposed based on diagnosis analysis of our experiment results and we name it as ``Twin-DQ'':
        \begin{equation}
        \begin{aligned}
                 R^{\rm{Twin-DQ}}(t) = & - \sum\limits_{j=0}^{11} (d_j^{t+1}(A_i^k)+q_j^{t+1}(A_i^k))  \\ & -  \sum\limits_{j=12}^{23} (d_j^{t+1}(A_i^k)+q_j^{t+1}(A_i^k)) -(d_j^t(A_i^k)+q_j^t(A_i^k))
         \end{aligned}
        \end{equation}
        The aim of the first term of the equation is to minimize the delay and queue length of the upstream lanes, while the purpose of the second term is to minimize the difference of DQ between timesteps $t$ and $t+1$. The basic idea of ``Twin-DQ'' is to consider the rewards of upstream lanes and downstream lanes separately. An action of the traffic signal can affect the queue length and delay in the upstream lanes directly. However, the influence for the downstream lanes is more complicated. Thus, the reward ``Twin-DQ'' tries to maintain the congestion level of the downstream lanes, which may facilitate the coordination between upstream and downstream intersections.
        In our experiments, choose $k$ equal to$100$, achieve best results in the local evaluation dataset.

\subsection{Control Scheme}\label{sec:control}
By default, the decision making process by the agent for a given intersection is triggered if the green time exceeds 30 (or 20) seconds. A major reason for this setting is that switching to another phase would result in a 5-second ``all red'' period, during which all vehicles may not pass this intersection. Thus, a default setting of a relatively long green time (e.g., 20 or 30 seconds) can avoid such a situation and guarantee higher returns in the long run.             
However, it is observed that there are situations in which the current phase goes on with no vehicles in upstream lanes. A more sophisticated approach is to consider not only the green time of the current
 phase, but also alternative conditions that may trigger the agent to make better decisions. As a result, the suite of control schemes we use to trigger the DQN-agent to do the phase selection action is as follows:
\begin{description}
    \item [Condition 1]: the green time of the current signal phase exceeds 30 seconds.
    \item [Condition 2]: the queue length of the current signal phase in the upstream zone of influence with $k=60$ equals zero. 
    \item [Condition 3]: the queue length of the current signal phase in the downstream zone of influence with $k=60$ equals 8.
    \item [Condition 4]: the queue pressure of the current signal phase in the zone of influence with $k=60$ is less than -5.
    
\end{description}
During the training process of the DQNs, if any of conditions above is met, the DQN-agent is triggered to begin the phase selection process. 

\subsection{DQN Framework and Training Scheme}

\begin{table}[!t]
    \centering
    \caption{Training parameters of DQN}
    \begin{center} \small
    \begin{tabular}{l l}
    \toprule
       Parameter & Value  \\ \midrule
        greenSec & 20 seconds \\ 
        Gamma &  0.8 \\ 
        Model update frequency & 1 \\  
        Target model update frequency & 17 \\
        Learning rate & 5e-5  \\
        epsilon  & 0.2 \\ 
        epsilonMin & 0.01 \\
        Loss function & Huber loss (smooth-L1) \\ \midrule
    \end{tabular}
    \end{center}
    \label{table:training}
\end{table}

Our training framework follows the classical paradigm of Double Q-Network (double DQN)~\cite{van2016deep,mnih2015human} which includes an online network and a target network.
The output dimensions of both networks are set to 8, which correspond to 8 candidate signal phases to be selected.
Meanwhile, We set the the parameter $\gamma$ to 0.8, which aims at maximizing the total rewards of the next 5 rounds.
The $\epsilon$-greedy policy is adopted to train the model.
A detailed description of the training parameters are presented in Table~\ref{table:training}.

To stabilize the training process of the DQN framework, we design a two-objective neural network. Specifically, one objective is used for predicting the Q-values, and the other one is used for predicting the rewards. 
The loss function consists of two terms, namely, the smooth-L1 loss of the predicted Q-values and the smooth-L1 loss of the predicted rewards.

As the reward signals are straightforward, it is observed that this approach can greatly accelerate and stabilize the convergence of the training process in our experiments. 

\subsection{Rule-based Agent}\label{sec:rule}
In this subsection, we describe a rule-based agent design in addition to the DQN agent. The core of the rule-based agent is the calculation of vehicle density near the center of the intersections. In order to alleviate traffic congestion in multiple directions, phases with high vehicle density on both its upstream lanes are given higher priority. For each upstream lane, the density is calculated as 
\begin{equation}
    \alpha^{\rm{up}}_j = \frac{x_j(A^{k_{\rm{up}}}_i)}{k_{\rm{up}}}
\end{equation}
where $k_{\rm{up}}$ is a global value for the zone of influence in the upstream lanes. The density of downstream lanes is calculated as 
\begin{equation}
    \alpha^{\rm{down}}_j = \frac{x_j(A^{l_j-k_{\rm{up}}}_i)}{l_j-k_{\rm{up}}}
\end{equation}
where $l_j$ is the length of the downstream lane with index $j$, such that the range of $l_j-k_{\rm{up}}$ is complement to the upstream zone of influence. The order of priority for each upstream lane, $o_{j_1}$, is decided by the relative density $\alpha_{j_1} = \alpha^{\rm{up}}_{j_1} - \mu \alpha^{\rm{down}}_{j_2}$, where $j_2$ is the index of the corresponding downstream lane for the upstream lane $j_1$ and $\mu$ is a coefficient to be tuned. Specifically, the values of order ranges from 1 to 12 (1 for the highest priority). The difficulty of the designing of the rules mainly comes from the fact that each phase generally correspond to two upstream-downstream lane pairs. An empirical observation is that a phase with two high-priority lane pairs is more favorable than a phase with one top-priority lane and a low-priority lane. Although designing a perfect chain of rules is unlikely, carefully tuning the requirement for both lanes of a phase to be balanced is crucial.

More specifically, the rule has 4 layers of logic. If none of the phases satisfy the condition of a layer, the logic flow moves on to the next layer. For a given phase being considered, we link the dominant upstream-downstream lane pair with index $j_{\mathbf{I}}$ (same as its upstream lane index), while the other lane pair is linked with index $j_{\mathbf{II}}$. The layers of logic are given as follows:
\begin{itemize}
\item \textbf{Layer 1}: the first layer aims at picking a phase that corresponds to a lane blocked for too long. The number of blocked round for each lane is calculated and ordered. At the first round, a lane pair $j_{\mathbf{I}}$ is selected if its upstream lane is blocked for more than $C_{\rm{block}}$ seconds and its blocked time is the largest. A phase will be selected if $o_{j_{\mathbf{I}}} <= 5$ and $o_{j_{\mathbf{II}}} <= 4$. If more than one phase satisfies the conditions, the phase serving two roads is picked. At the second round, a lane pair $j_{\mathbf{I}}$ is chosen if its upstream lane is blocked for more than $C_{\rm{block}} + 50$ seconds and its blocked time is the second largest. A phase will be selected if $o_{j_{\mathbf{I}}} <= 2$ and $o_{j_{\mathbf{II}}} = 1$. The value of $C_{\rm{block}}$ starts with 200 and increases over time until it reaches 300.

\item \textbf{Layer 2}: the second layer aims at picking a phase that corresponds to two high-density lanes. At this layer, the vehicle densities for both lane pairs of each phase are checked and a phase will be selected if it has two balanced, high-relative-density lane pairs. Given the balance coefficient $C_{\rm{balance}}$, a lane pair is balanced if $\alpha_{j_{\mathbf{I}}} < \alpha_{j_{\mathbf{II}}} / C_{\rm{balance}}$ (suppose $\alpha_{j_{\mathbf{I}}} >= \alpha_{j_{\mathbf{II}}}$). The selection of the second layer has a total of 5 rounds. For the first 4 rounds, the lane pair with $o_{j_{\mathbf{I}}} = 1$ is picked in advance, and the lane pair $j_{\mathbf{II}}$ can be selected if the two lane pairs are balanced and $o_{j_{\mathbf{II}}}$ is less than or equal to 2, 3, 4, or 5, respectively. For the 5th round, a phase is selected if its two lane pairs satisfy $o_{j_{\mathbf{I}}} <= 2$ and $o_{j_{\mathbf{II}}} <= 3$ and the two pairs are balanced. The values of $C_{\rm{balance}}$ for the 5 rounds start with 0.15, 0.2, 0.2, 0.2, and 0.25 and they decrease over time until they reach 0.13, 0.18, 0.18, 0.18, and 0.23, respectively.

\item \textbf{Layer 3}: the third layer aims at picking a phase that corresponds to two low-speed lanes. Here, the threshold for low speed, $C_{\rm{speed}}$, is set to 1 m/s. The average speeds of the upstream lanes are calculated as $s_j$, and the speeds are sorted in ascending order (denoted as $o^{\rm{s}}_j$). The prerequisite for a phase to be selected is that the average speeds at the two upstream lanes corresponding to the phase are lower than $C_{\rm{speed}}$. In addition, at the first round, a phase is selected if $o^{\rm{s}}_{j_{\mathbf{I}}} <= 2$, $o^{\rm{s}}_{j_{\mathbf{II}}} <= 2$, $o_{j_{\mathbf{I}}} <= 4$, and $o_{j_{\mathbf{II}}} <= 4$. Further, at the second round, a phase will be selected if $o^{\rm{s}}_{j_{\mathbf{I}}} <= 3$, $o^{\rm{s}}_{j_{\mathbf{II}}} <= 3$, $o_{j_{\mathbf{I}}} <= 5$, and $o_{j_{\mathbf{II}}} <= 5$. 

\item \textbf{Layer 4}: the final layer picks the phase corresponding to the lane with the highest density if none of the previous conditions can be satisfied. At the same time, a phase that serves two roads is given higher priority (unless one of the roads does not exist). A second round of phase selection is added if the lane with the highest density is a right-turning lane.
\end{itemize}

\section{Experiments}\label{sec:experiment}
In this section, we first show the performance of three different rewards, namely,  ``DQ'', ``Pressure'', and "Twin-DQ", under difference control schemes on the default round3\_flow0 traffic data. We then present our final experimental results in the leader-board.

The evaluation metrics we use here are the total number of vehicles served and the delay index.
The performance of the agent is evaluated every 20 seconds, and the evaluation process is terminated when the average delay index reaches the predefined threshold 1.40.

To show the effectiveness of the control scheme, we test
several different triggering policies (TP) during the evaluation stage:

\begin{enumerate}
    \item TP1: TP1 is the same as condition 1 described in subsection~\ref{sec:control}: the green time of the current signal phase exceeds 30 seconds.
    \item TP2: TP2 includes the condition 1 and condition 2 described in subsection~\ref{sec:control}. If one of these two conditions above is met, the DQN-agent is triggered to begin the phase selection process.
    \item TP3: TP3 includes all conditions described in subsection~\ref{sec:control}, if any of these conditions is met, the DQN-agent is triggered to begin the phase selection process.
\end{enumerate}
\begin{table}
    \caption{Evaluation results of number of vehicles served and delay index of different reward function-control scheme pairs on the default round3\_flow0 traffic data.
} 
    \begin{center}
        \begin{tabular}{lrrrr}
            \toprule
            Rewards & TP1 & TP2 & TP3   \\ \midrule
            DQ & 46,619/1.400 & 48,201/1.409 & 48,201/1.403 \\
            Pressure &47,747/1.403  & 48,201/1.402  & 48,201/1.400\\
            Twin-DQ & 47,747/1.400 & 48,201/1.400& 48,590/1.418 \\ \midrule
    \end{tabular}
    \end{center}
    \label{Tab:expcontrol}
\end{table}
The evaluation results of reward function-triggering policy pairs are illustrated in Table~\ref{Tab:expcontrol}. Briefly speaking, The ``Twin-DQ'' reward together with TP3 achieves the best evaluation score. Note that TP3 corresponds to the entire set of triggering conditions used for training the DQNs.

Further, Table~\ref{Tab:exp} illustrates our experiment results from the leader-board. Specifically, a single DQN model can serve about 313,035 vehicles. The rule-based agent described in subsection~\ref{sec:rule} also serves 313,035 vehicles with a delay of 1.402. A straightforward ensemble approach, which averages the predicted Q-values of multiple DQN models, can improve the number of vehicles served in the leader-board to 314,467 with an ensemble of 10 models. Our best score is based on the DQN with rule revising (adopting the decisions of the rule-based agent in a small set of cases), which can serve 317,954 cars with a delay of 1.401.


\begin{table}
    \caption{Experiment results in the leader-board. The ``Twin-DQ'' reward function is used to train the model.}
    \begin{center}
        \begin{tabular}{lcc}
            \toprule
            Strategy & Number of vehicles served & Delay  index \\ \midrule
               Rule-based & 313,035 & 1.402 \\
            DQN-single & 313,035 & 1.403 \\
            DQN-ensemble &314,467  & 1.404 \\
            DQN+Rule-based & 317,954 &1.401 \\ \midrule
    \end{tabular}
    \end{center}
    \label{Tab:exp}
\end{table}

\section{Conclusion}
In this paper, we present a DQN-based solution for the ``City Brain Challenge'' competition.
We described our overall analysis and details of the DQN-based framework for real-time traffic signal control.
Our main improvements on the leader-board come from two points: a newly-proposed reward function, namely, ``Twin-DQ'', and a suite of control schemes.
Meanwhile, an ensemble of multiple DQN models can further improve the performance.
In addition, a drawback of the DQN-based control framework in practice is that it may fail to make good decisions in certain cases due to the complexity of real-world road networks and traffic flow patterns. Therefore, applying heuristic rules to revise the DQN control actions in these cases can improve the performance of the DQN-based control framework.
The codes of our solution is available from \url{https://github.com/oneday88/kddcup2021CBCBingo}. Our work could serve, to some extent, as a baseline solution to traffic signal control of real-world road network  and inspire further attempts and researches.

\bibliographystyle{ACM-Reference-Format}
\bibliography{ref}

\appendix
\end{document}